%% file: main.tex
\documentclass{SCIS2024}

\def\eg{\emph{e.g.}}
\def\etc{\emph{etc}}

\usepackage{tcolorbox}
\usepackage{xspace}
\usepackage{subcaption} 
\usepackage{bbding}
\usepackage{booktabs}
\usepackage{makecell} 
\usepackage{color, colortbl} 
\definecolor{Gray}{gray}{0.95}
\usepackage{tabularx}
\usepackage{wrapfig}
\usepackage{multirow}
\usepackage{latexsym}

\definecolor{darkgreen}{rgb}{0.259, 0.537, 0.145}

\newcommand{\dataname}{\textsc{MMInstruct}\xspace}

\begin{document}
\ArticleType{RESEARCH PAPER}
\Year{2024}
\Month{}
\Vol{}
\No{}
\DOI{}
\ArtNo{}
\ReceiveDate{}
\ReviseDate{}
\AcceptDate{}
\OnlineDate{}

\title{MMInstruct: A High-Quality Multi-Modal Instruction Tuning Dataset with Extensive Diversity}{Title keyword 5 for citation Title for citation Title for citation}

\author[4$\dagger$]{Yangzhou LIU}{}
\author[4$\dagger$]{Yue CAO}{}
\author[7,1$\dagger$]{Zhangwei GAO}{}
\author[5,1$\dagger$]{Weiyun WANG}{}
\author[4,1$\dagger$]{Zhe CHEN}{}

\author[6,1$\dagger$]{\\Wenhai WANG}{}
\author[2]{Hao TIAN}{}
\author[2]{Lewei LU}{}
\author[3,1,2]{Xizhou ZHU}{}
\author[4]{Tong LU}{}
\author[1]{\\Yu QIAO}{}
\author[3,1]{Jifeng DAI}{daijifeng@tsinghua.edu.cn}

\AuthorMark{Liu Y Z}

\AuthorCitation{Yangzhou LIU, Yue CAO, Zhangwei GAO, et al}

\address[1]{Shanghai AI Laboratory, Shanghai {\rm 200232}, China}
\address[2]{SenseTime Research, Shanghai {\rm 200233}, China}
\address[3]{Tsinghua University, Beijing {\rm 100084}, China}
\address[4]{Nanjing University, Nanjing {\rm 210023}, China}
\address[5]{Fudan University, Shanghai {\rm 200433}, China}

\address[6]{The Chinese University of Hong Kong, Hong Kong {\rm 999077}, China}
\address[7]{Shanghai Jiao Tong University, Shanghai {\rm 200240}, China}

\input{sections/0_abstract}


\maketitle

\input{sections/1_introduction}
\input{sections/2_related_work}
\input{sections/3_method}

\input{sections/4_experiments}

\input{sections/5_conclusion}

\newpage

\bibliographystyle{plain}
\bibliography{egbib}

\end{document}

%% file: sections/0_abstract.tex
\abstract{Despite the effectiveness of vision-language supervised fine-tuning in enhancing the performance of Vision Large Language Models (VLLMs). 
However, existing visual instruction tuning datasets include the following limitations: 
(1) Instruction annotation quality: despite existing VLLMs exhibiting strong performance, instructions generated by those advanced VLLMs may still suffer from inaccuracies, such as hallucinations.
(2) Instructions and image diversity: the limited range of instruction types and the lack of diversity in image data may impact the model's ability to generate diversified and closer to real-world scenarios outputs.
To address these challenges, we construct a high-quality, diverse visual instruction tuning dataset \dataname, which consists of 973K instructions from 24 domains. There are four instruction types: Judgement, Multiple-Choice, Long Visual Question Answering and Short Visual Question Answering.
To construct \dataname, we propose an instruction generation data engine that leverages GPT-4V, GPT-3.5, and manual correction. Our instruction generation engine enables semi-automatic, low-cost, and multi-domain instruction generation at 1/6 the cost of manual construction. Through extensive experiment validation and ablation experiments, we demonstrate that \dataname could significantly improve the performance of VLLMs, e.g., the model fine-tuning on \dataname achieves new state-of-the-art performance on 10 out of 12 benchmarks. 
The code and data shall be available at \url{https://github.com/yuecao0119/MMInstruct}.
}

%% file: sections/1_introduction.tex
\section{Introduction}

Benefiting from the large-scale parameters and extensive pre-training corpus, Large Language Models (LLMs)~\cite{chatgpt,touvron2023llama,2023internlm,touvron2023llama2,chiang2023vicuna} have demonstrated a range of powerful capabilities, including language generation, in-context learning, world knowledge, and commonsense reasoning.
Beyond the pre-training phase, these models undergo an additional training stage, termed instruction tuning, which equips these base models with the ability to follow user instructions, thus transforming them into chat models.
By integrating these chat models with pre-trained vision foundation models through a vision-language connector, Vision Large Language Models (VLLMs) exhibit impressive performance across various vision-language tasks. These models employ similar training schemes to empower VLLMs to effectively understand visual information.
Specifically, during the pre-training phase, models are trained to predict the next text token conditioned on a given image, while during the instruction tuning stage, the models are required to learn to interact with users conditioned on the given image and instructions.

However, existing multi-modal instruction tuning datasets~\cite{liu2023llava_1_5,wang2024allseeing_v2,chen2023sharegpt4v} include following limitations:
(1) \textbf{Image Diversity}: The images of these datasets are sourced from existing datasets, such as COCO~\cite{lin2014microsoft}, which is restricted to the common scenes and thus limits the models' generalization ability. For instance, models struggle to process the text-oriented OCR image.
(2) \textbf{Annotation Quality}: These datasets are generated automatically by employing models (\eg, GPT-4V~\cite{gpt4v}) to generate new question-answer pairs based on annotations from existing datasets. Despite the advanced capabilities of existing VLLMs, such data generation pipelines inevitably introduce noise to the generated dataset, leading to hallucinations in models.
(3) \textbf{Instruction Diversity}: The instruction types within these datasets are limited, negatively impacting the models' ability to generalize across the diverse range of real-world instructions.

\begin{figure}[t!]
    \centering
    \includegraphics[width=\textwidth]{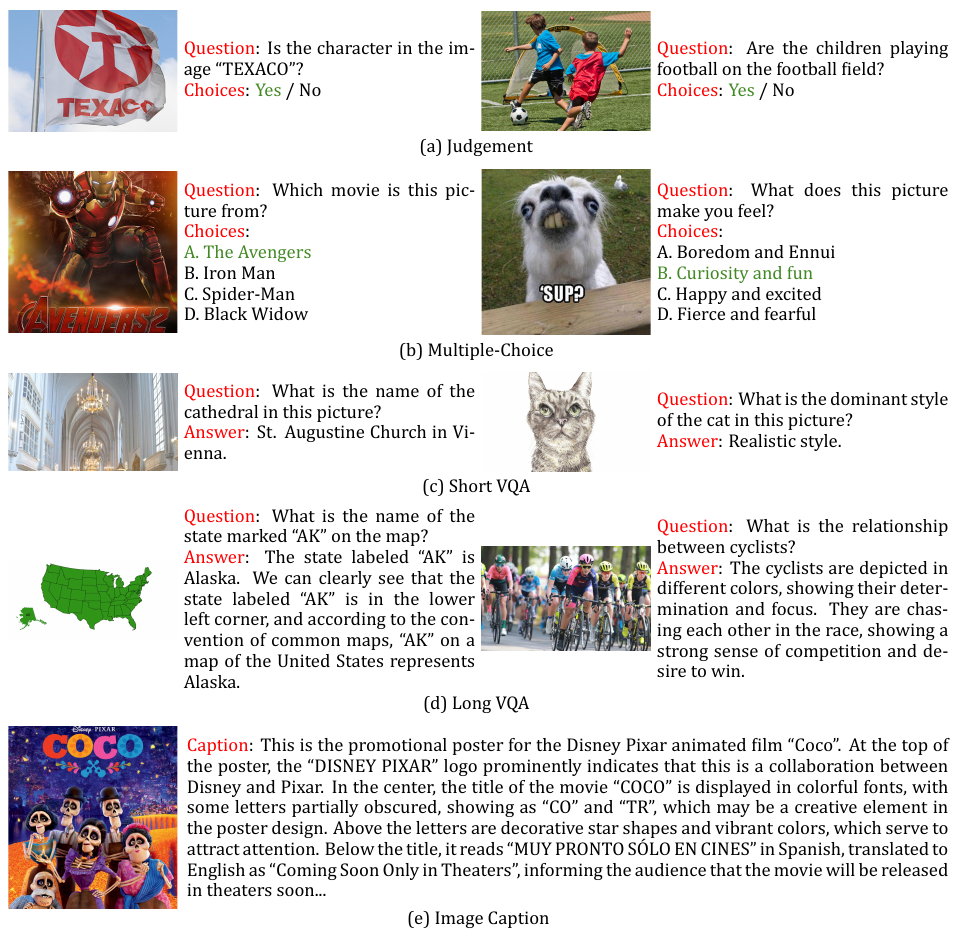}
    \caption{{Examples of various question types in \dataname.} (a) to (e) represent different question types in \dataname. \textbf{Question} denotes instruction generated by GPT. \textbf{Answer} denotes the response based on the instruction. \textbf{Caption} denotes the detailed image description generated by GPT. The green option indicates the correct answer.}
    \label{fig:example-in-question-type}
\end{figure}

\begin{figure}
    \centering
    \includegraphics[width=\textwidth]{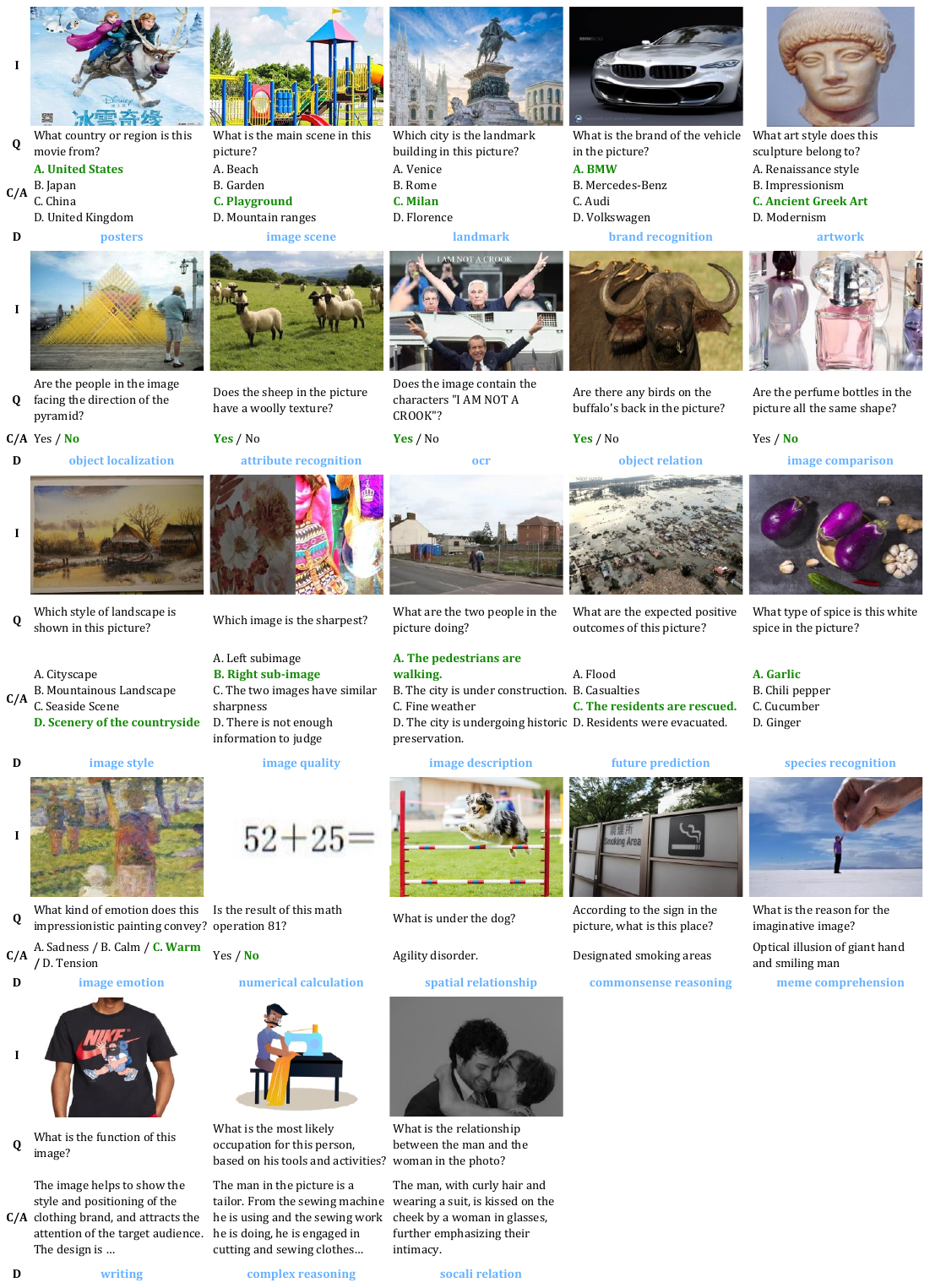}
    \caption{{Examples of different domains in \dataname.} \textbf{I} denotes Image. \textbf{Q} denotes instruction generated by GPT. \textbf{C/A} denotes options and the correct answer to the related instruction; in judgment and multiple-choice questions, the green option indicates the correct answer. \textbf{D} denotes the domain.}
    \label{fig:enter-label}
\end{figure}

To address these issues, we propose a high-quality and diverse visual instruction tuning dataset named {\dataname}, which contains 973K instructions. To achieve the universality of the dataset, we design 24 task domains commonly seen in daily life, including (1) Perception (image style, image scene, image quality, image comparison, object localization, object relation, attribute recognition, image description, OCR, posters, artwork, landmark, spatial relationship, brand recognition, species recognition); (2) Reasoning (numerical calculation, image emotion, commonsense reasoning, complex reasoning, social relation, future prediction, meme comprehension, writing); (3) Multi-Round Long Visual Question Answering (Multi-Round Long VQA).
We show some example instructions of various question types in Figure \ref{fig:example-in-question-type} and different domains in Figure \ref{fig:enter-label}. 
Specifically, our instructions comprise four common types: Judgement, Multiple-Choice, Long Visual Question Answering (Long VQA), and Short Visual Question Answering (Short VQA). The instructions do not adhere to a fixed template, and there may be variations in format among instructions with the same questioning purpose. And we additionally construct Multi-Round Long VQA data for long-context logical reasoning training of the model.

Relying on manual efforts to construct such a diverse and high-quality dataset can be excessively expensive, especially when the data scale is large.
Therefore, we propose a semi-automatic, low-cost instruction generation data engine that leverages GPT-4V~\cite{gpt4v}, GPT-3.5~\cite{chatgpt}, and manual correction. To enrich the scope of image coverage, we first utilize web crawlers and similarity searches to swiftly gather a large quantity of high-quality images. Then, these images undergo deep semantic analysis via GPT-4V, transcending the mere reliance on rudimentary annotations of the images themselves. After that, we integrate the characteristics of both the images and domains to design approximately ten seed questions for each domain. Unlike other datasets, the seed questions in our engine serve merely as references, encouraging GPT to generate diverse forms of instructions. Specifically, questions and answers are generated at the same time to ensure accuracy and reduce illusions. 
In this way, the data engine can automatically generate detailed semantic captions and diverse instructions for the image.  Additionally, manual corrections are integral throughout the entire process to ensure dataset quality and minimize biases.

\begin{table}[t]
\fontsize{8pt}{10pt}\selectfont
 \centering
    \caption{{Comparison of \dataname with existing visual instruction tuning dataset.} Note that we unify the division of instruction tasks for all datasets based on our task domain partitioning. Question types are abbreviated due to space constraints. TF: judgment; MC: multiple-choice; LVQA: Long VQA; SVQA: Short VQA.}
    \begin{tabular}{p{3cm} | >{\centering\arraybackslash}p{2.1cm} >{\centering\arraybackslash}p{2.1cm} >{\centering\arraybackslash}p{3.4cm} >{\centering\arraybackslash}p{3.2cm} }
    
    \toprule
    {Dataset} & {\#Instances}  & {\#Domains}  & {Question Types} & {Question Form} \\ %
    \midrule 
    LLaVA~\cite{liu2023llava_1_5} & 150K & 3 & LVQA, SVQA & Fixed   \\ 
    ShareGPT4V~\cite{chen2023sharegpt4v} & 100K & 1 & LVQA & Fixed  \\  
    M\textsuperscript{3}IT~\cite{li2023m} & 2.4M & 12 & TF, MC, LVQA & Fixed \\ 
    Shikra~\cite{chen2023shikra} & 156K & 10 & LVQA & Diverse  \\ 
    InstructBLIP~\cite{instructblip} & 1.6M & 12 & TF, MC, LVQA & Fixed  \\ 
    MultiInstruct~\cite{xu2022multiinstruct} & 510K & 14 & TF, MC, LVQA & Diverse  \\
    Vision-Flan~\cite{xu2024vision} & 1.6M & 22 & TF, MC, LVQA & Fixed  \\ 
    \rowcolor{Gray}
    \dataname~(Ours) & 973K & 24 & TF, MC, LVQA, SVQA & Diverse  \\ 
    \bottomrule
    \end{tabular}
    \label{tab:compare}
\end{table}

\begin{wraptable}{h}{0.443\textwidth}
    \vspace{-0.48cm} 
   \fontsize{8pt}{10pt}\selectfont
    \centering
    \caption{{Comparison of costs between \dataname construction and manual construction.} \textbf{Total} refers to the estimated cost of building the \dataname.}
    \begin{tabular}{l|cc}
    \toprule
    {Method} & \makecell[c]{{Manual} \\ {Construction}} & {\dataname} \\
    \midrule 
    Per Image  & - & \$0.00885  \\
    Per Instruction  & \$0.84 & \$0.0004 \\
    Total & \$817, 320  & \$128, 304 \\
    \bottomrule
    \end{tabular}
    \label{tab:cost} 
\end{wraptable}

As shown in Table \ref{tab:compare}, we compare \dataname with some representative visual instruction tuning datasets, demonstrating significant advantages in terms of coverage and diversity of our instructions. Furthermore, when compared to the exclusive dependence on manual dataset construction, our data engine's cost is only 1/6 of manual annotation while concurrently ensuring data quality. The cost comparison between manual construction and \dataname is shown in Table \ref{tab:cost}.

\begin{figure}[h!]
    \centering
    \includegraphics[width=0.88\textwidth]{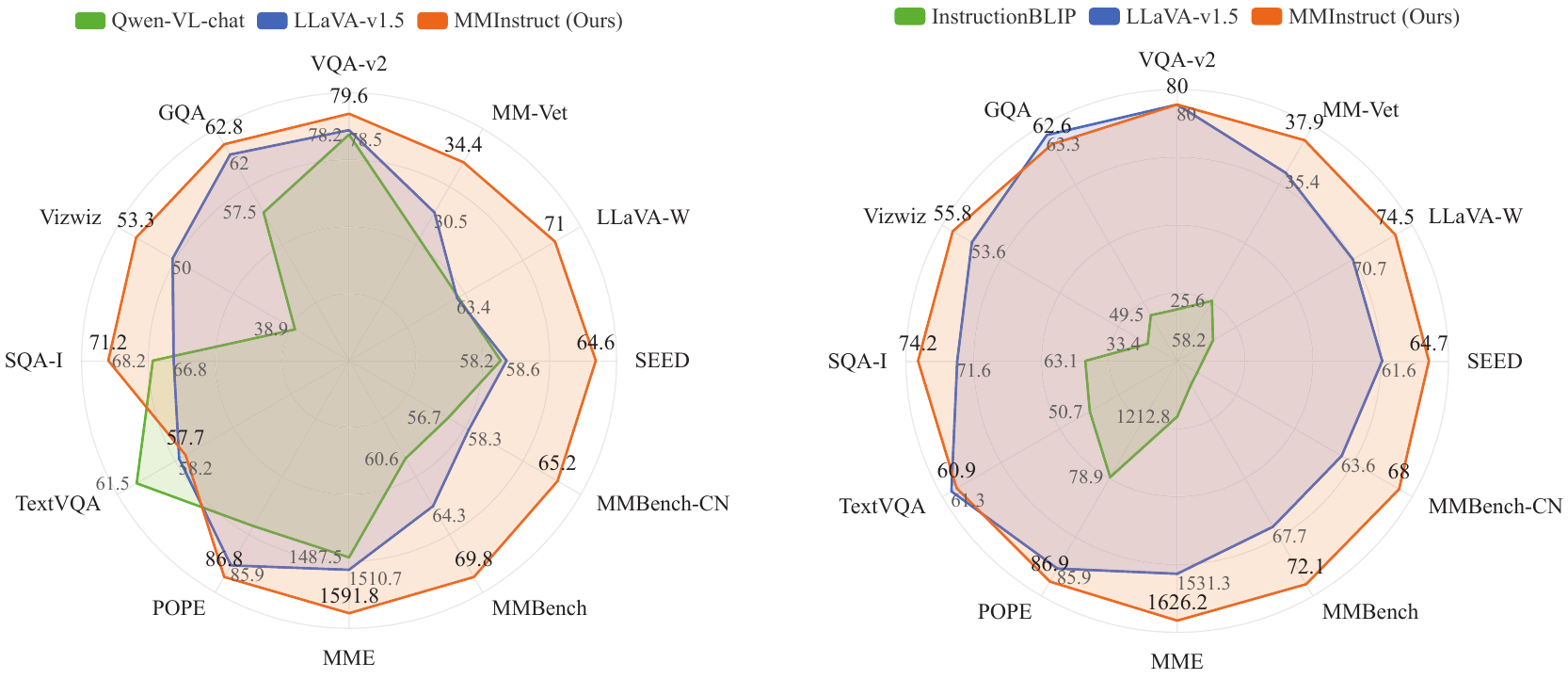}
    \caption{
    {Performance comparison of different model sizes.} (a) Compared with 7B models including Qwen-VL-Chat~\cite{bai2023qwenvl}, LLaVA-1.5-7B~\cite{liu2023llava_1_5}, our model achieves SoTA on 11 benchmarks. (b) Compared with 13B models, including InstructBLIP~\cite{instructblip}, LLaVA-1.5-13B~\cite{liu2023llava_1_5}, our model achieves SoTA on 10 benchmarks.
    }
    \label{fig:results-benchmarks}
\end{figure}

To verify the effectiveness of \dataname, we incorporate \dataname into the instruction fine-tuning phase of LLaVA-1.5\cite{liu2023llava_1_5}. Our experimental results demonstrate that \dataname significantly enhances the capabilities of VLLMs. 
Figure \ref{fig:results-benchmarks} shows the performance comparison of LLaVA-1.5~\cite{liu2023llava_1_5} on different benchmarks after fine-tuning on LLaVA-665K and \dataname. We can see that after fine-tuning on \dataname, our model demonstrates impressive improvements across a wide range of evaluation benchmarks and exceeds LLaVA-1.5 on 10 out of 12 benchmarks.
We also conduct extensive ablation experiments to analyze the impacts of varying the fine-tuning data on VLLMs.
These results highlight the effectiveness of \dataname.

In conclusion, our paper makes the following contributions:
\begin{itemize}
    \item  We construct a visual instruction tuning dataset \dataname, containing 24 common domains. \dataname comprises 973K high-quality and diverse visual instructions featuring diverse question forms and types, including judgment, multiple-choice, Long VQA, and Short VQA.
    \item To construct \dataname, we designed a semi-automatic, low-cost instruction generation data engine based on GPT-4V, GPT-3.5, and manual correction. 
    Compared with purely manual construction, Our data engine's cost is only 1/6 of manual annotation while ensuring annotation quality and data diversity.
    \item We conduct comprehension experiments to validate the effectiveness of \dataname. As shown in Figure \ref{fig:results-benchmarks}, after fine-tuning on \dataname, LLaVA-1.5 achieves state-of-the-art results on 10 out of 12 benchmarks. Specifically, the scores on MME~\cite{fu2023mme} and LLaVA-Bench (In-the-Wild)~\cite{liu2023llava} are 1626.2 and 74.5, surpassing LLaVA-1.5 by 94.9 and 3.8 points respectively. 
\end{itemize}

%% file: sections/2_related_work.tex
\section{Related Work}
\paragraph{Vision Large Language Models.} 
Significant progress has been achieved in the realm of Vision Large Language Models (VLLMs). 
Models like CLIP~\cite{radford2021clip}, ALIGN~\cite{jia2021align}, EVA~\cite{fang2022eva},  which are trained via contrastive learning-based methods, demonstrate the capacity to understand the complex semantics of the open-world through image-text alignment.
Subsequent endeavors, as exemplified by VL-BERT~\cite{su2019vl_bert}, VL-BEiT~\cite{bao2022vlbeit}, ALBEF~\cite{li2021albef}, VLMo~\cite{vlmo}, BEiT-3~\cite{wang2023beit3}, CoCa~\cite{yu2022coca}, and the Uni-Perceiver series~\cite{zhu2021uni, zhu2022uni, li2023uni}, have shown proficiency in performing a variety of multi-modal downstream tasks.
However, these models are trained from scratch, leading to escalated expenses in the development of novel models.

In recent years, numerous VLLMs~\cite{alayrac2022flamingo,chen2023internvl,chen2024internvl_v1_5,liu2023interngpt,chen2022pali,lai2023lisa,wang2023allseeing,tian2024mminterleaved,wu2024visionllmv2,dong2024ixc4khd} have been developed by incorporating pre-trained LLMs~\cite{touvron2023llama,touvron2023llama2,chiang2023vicuna,2023internlm} with off-the-shelf vision encoders (\eg, CLIP~\cite{radford2021clip}), aiming to combine the visual encoding capabilities of vision encoders alongside the linguistic knowledge of language models. 
Earlier research, such as Frozen~\cite{tsimpoukelli2021multimodal} and VisualGPT~\cite{chen2022visualgpt}, demonstrates the efficiency of employing LLMs as decoders for VLLMs, facilitating learning from multi-modal data. 
Flamingo~\cite{alayrac2022flamingo} can employ interleaved texts and images as input and is endowed with remarkable few-shot learning capabilities. 
In VLLMs, a feature resampler~\cite{li2023blip2} or projection layer~\cite{liu2023llava,liu2023llava_1_5} is employed to align the features of vision encoders with the embedding space of language modes. This alignment facilitates the LLM in acquiring the ability to understand images. 
With the introduction of visual instruction tuning in VLLMs (\eg, InstructBLIP~\cite{instructblip}, Qwen-VL~\cite{bai2023qwenvl}, InternVL~\cite{chen2023internvl,chen2024internvl_v1_5}, GPT-4~\cite{gpt4v}, LLaVA series~\cite{liu2023llava,liu2023llava_1_5}, Gemini series~\cite{team2023gemini,reid2024gemini_1_5}), a significant enhancement has been observed in the capability to follow instructions and complete visual tasks. However, many advanced models~\cite{gpt4v, bai2023qwenvl} do not publish their SFT datasets. Currently, the community urgently needs a diverse, high-quality, open-source visual instruction dataset to further improve model performance.

\paragraph{Datasets for Vision-Language Supervised Fine-tuning.} 
In the NLP community, the utilization of instruction-following data~\cite{ouyang2022training, wang2022self, wang2022benchmarking} during the SFT stage enables Large Language Models (LLMs) to acquire the capability of following natural language instructions and solving real-world tasks, thus contributing to notable advancements~\cite{openai2022chatgpt,openai2023gpt4,chiang2023vicuna,touvron2023llama2}. The integration of the vision modality further enhances this process by providing additional information for interactions, making visual instruction tuning a more creative and innovative procedure.

MultiInstruct \cite{xu2022multiinstruct} introduces the first human-label visual instruction tuning dataset. Mini-GPT4 \cite{zhu2023minigpt4} generated its instruction-based dataset by composing image-text datasets and handwritten instruction templates. LLaVA\cite{liu2023llava} employs ChatGPT/GPT-4 to convert image-text pairs into multi-modal instruction-following data. Subsequently, several instruction datasets (\eg, LAMM \cite{yin2024lamm}, MIMIC-IT \cite{li2023mimic}, and Macaw-LLM \cite{lyu2023macaw}) further encompass 3D-domain, Audio and videos examples for instruction tuning. InstructBLIP \cite{instructblip} and LLaVA-1.5\cite{liu2023llava_1_5} incorporate academic-task-oriented Visual Question Answering (VQA) datasets to augment the model’s visual capabilities. M$^3$IT\cite{li2023m} further scaled up the instruction data to 2.4 million instances.

Some works focus on improving the performance of VLLMs in specific domains or emphasize enhancing certain aspects of the model's capabilities. VideoChat\cite{2023videochat}, TimeChat\cite{Ren2023TimeChat}, and Valley\cite{luo2023valley} build video-centric instruction datasets aimed at enhancing the video comprehension, conversation, and complex reasoning capabilities of VLLMs. ScienceQA\cite{lu2022sqa} and MMMU\cite{yue2023mmmu} construct question-answer pairs from primary and secondary school classes and college exams, respectively, covering diverse disciplines and emphasizing perception and reasoning with domain-specific knowledge. LLaVAR \cite{zhang2023llavar} augments visual instruction tuning with text-rich images using OCR tools and GPT-4. Some datasets (\eg, mPLUG-DocOwl\cite{ye2023mplugdocowl}, InstructDoc\cite{InstructDoc2024}) focus on document understanding tasks, necessitating models to possess robust OCR capabilities as a foundation. LRV-Instruction \cite{liu2023lrv} includes both positive and negative instructions to mitigate hallucination, resulting in a more robust model. 
Shikra \cite{chen2023shikra} and All-Seeing\cite{wang2023allseeing,wang2024allseeing_v2} utilize data with region annotations to enhance the referential dialogue and panoptic visual recognition and understanding capabilities of VLLMs. Vision-Flan \cite{xu2024vision}, consisting of 22 tasks drawn from academic datasets, is built to address issues of poor generalization, hallucination, and catastrophic forgetting in models trained on GPT-4 synthesized data.

Compared to the previous works, we aim to construct a visual instruction tuning dataset that encompasses a wider range of domains, features more precise annotations, and provides richer question-answering forms and types.

%% file: sections/3_method.tex
\section{Method}

\begin{table}[t!]
\fontsize{8pt}{10pt}\selectfont
 \centering
    \caption{{Domain partitioning details of \dataname.} It includes 23 single turns and 1 multi-round long visual question answering.}
    \begin{tabular}{p{2cm}|p{13cm}}
    \toprule
    {Conv Type} & {Domains} \\ 
    \midrule 
    \makecell[l]{Single-Turn \\ (Perception)} & \makecell[l]{image style, image scene, image quality, image comparison, object localization, object relation, \\ attribute recognition, image description, OCR, posters, artwork, landmark, spatial relationship, \\ brand recognition, species recognition} \\
    \midrule 
    Single-turn (Reasoning) & numerical calculation, image emotion, commonsense reasoning, complex reasoning, social relation, future prediction, meme comprehension, writing \\
    \midrule 
    Multi-Round & multi-round long visual question answering \\
    \bottomrule
    \end{tabular}
    \label{tab:domain-list} 
\end{table}

In this paper, we propose a visual instruction tuning dataset, named \dataname, which ensures diverse images, high annotation quality, and diverse instructions. Our dataset is primarily divided into 24 domains, including 23 single-turn question-answering domains and one multi-round long visual question-answering domain. The partitioning details are shown in Table \ref{tab:domain-list}. \dataname comprises a total of 161K high-quality detailed image captions and 973K instruction data. 

\begin{figure}[t!]
    \centering
    \includegraphics[width=\textwidth]{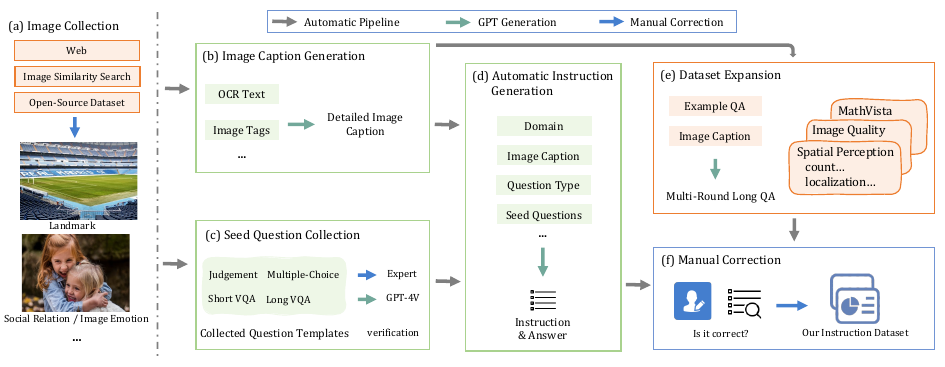}
    \caption{{Data engine for \dataname.} Our data engine consists of automatic generation and manual correction. (a) We collect a large number and diversity of images from a variety of sources. (b) We utilize GPT-4V to generate detailed image descriptions based on the image and context of the image. (c) Human experts collect seed questions and validate the effectiveness of seed questions using GPT-4V. (d) Then, leveraging those detailed image descriptions and seed questions, we employ GPT-3.5 to generate Instruction-Answer pairs. (e) We also use several methods to expand our dataset. (f) Finally, additional manual corrections are performed.}
    \label{fig:pipeline}
\end{figure}

To overcome the high cost of dataset construction while increasing dataset coverage and diversity, we propose a semi-automatic and low-cost instruction generation data engine utilizing GPT-4V, GPT-3.5 and manual correction, as shown in Figure \ref{fig:pipeline}. Our data engine comprises six steps: (a) Image Collection, (b) Image Caption Generation, (c) Seed Question Collection, (d) Automatic Instruction Generation, (e) Dataset Expansion, and (f) Manual Correction. Initially, we collect a large and diverse set of images from various sources and employ GPT-4V to generate detailed image captions. Seed questions are curated by our experts and validated for effectiveness. Subsequently, leveraging both the image captions and seed questions, GPT-3.5 automatically generates a rich and diverse set of instruction data. Additionally, we employ various methods to expand our dataset. Finally, manual corrections are made to ensure data quality and accuracy.

Our efforts primarily revolve around the following: (1) \textbf{Image Diversity}: Since high-quality images are difficult to obtain, image acquisition in existing instruction datasets mostly relies on open-source image datasets, but this also limits the scope of image inclusion. Therefore, we propose a process to quickly and extensively collect images from the Internet according to a specific domain, with multiple manual screenings to ensure image quality, as introduced in Section \ref{sec:img_col}.
(2) \textbf{Annotation Quality}: Existing datasets typically rely on existing annotations of images for instruction generation. Rough annotations can cause hallucination problems. Therefore, in Section \ref{sec:img_cap}, we propose leveraging GPT-4V to obtain rich semantic information from images, followed by manual corrections to ensure annotation quality.
(3) \textbf{Instruction Diversity}: For a specific domain, users have various instructions. In Section \ref{sec:seed_ques}, we propose compiling a seed question set by analyzing common user instructions. For each image, diverse instructions of four types are generated using the generation pipeline outlined in Section \ref{sec:inst_gen}. Additionally, in Section \ref{sec:data_expan}, we generate multi-round Long VQA instructions and incorporate other open-source datasets to further supplement our dataset.

\subsection{Image Collection} \label{sec:img_col}

\begin{table}[t!]
\fontsize{8pt}{10pt}\selectfont
 \centering
    \caption{{Some key phrases used for searching images on the web.}}
    \begin{tabular}{p{3.5cm}|p{11.6cm}}
    \toprule
    {Domains} & {Search key phrases} \\ 
    \midrule 
    Feature Prediction & typhoon, traffic accident, football match, dance, rocket launch, sunrise... \\
    Species Recognition & mammals, marine life, reptiles, insect, virus, plants, fruits...\\
    Meme Comprehension & pepe the frog, confession bear meme, bad luck brian, doge meme... \\
    \bottomrule
    \end{tabular}
    \label{tab:search-keyword} 
\end{table}

In order to effectively reduce costs while ensuring image diversification, we propose an efficient image collection process. Firstly, experts define key phrases for each domain 
based on which a large number of open images are crawled from the web, and preliminary screening is conducted on the crawled results. In Table \ref{tab:search-keyword}, a small portion of the utilized search key phrases is listed. Next, leveraging existing images as a foundation, by utilizing $k$-Nearest Neighbors image similarity search in the large-scale image repository Laion-5B~\cite{schuhmann2022laion5b}, rigorously assessing their suitability and quality. 
Note that in order to avoid duplication of images, we strictly deduplicate them through image annotation information and manual screening.
Finally, we organize the existing images and select some images from open-source datasets as a final supplement. Through this approach, we collect a total of 161,000 high-quality images across the 24 domains.

\subsection{Image Caption Generation} \label{sec:img_cap}
Previous research \cite{liu2023llava, zhang2023llavar, liu2023lrv} has relied on annotation data, including scene captions, bounding boxes, and OCR, to depict images and generate instructions for text-only GPT models. However, those poor annotations have become a bottleneck in generating instructions. Therefore, in our data engine, we use GPT-4V to generate a sufficiently detailed and domain-specific image caption for each image. Rich image information can make instruction generation more diverse and reduce hallucination problems.

It is worth noting that we have also implemented the following measures to ensure the accuracy and coherence of the captions generated by GPT-4V. In our caption generation prompt, we additionally integrate domain-specific prior knowledge. For example, in the OCR domain, text recognition is initially performed on images using Google OCR. And we modify the prompt appropriately for different domains. 
In addition, the annotations of the collected images themselves are also integrated into the prompt for caption generation. The fundamental prompt for all domains and the addition prompt added for specific domains are listed in Table \ref{tab:caption-prompt}.

\begin{table}[t!]
\fontsize{8pt}{10pt}\selectfont
 \centering
    \caption{{Prompt used for detailed image description generation.} \textbf{Universal} represents the fundamental prompt applicable to all domains. The remaining lines enumerate additional prompt content added for specific domains. \textit{(if has)} means the content is applicable to the corresponding situation.}
    \begin{tabular}{p{3.2cm}|p{11.9cm}}
    \toprule
    {Domains} & {Prompt} \\ 
    \midrule 
    \textbf{Universal}\newline(For all domains) & \makecell[l]{$<$Image$>$ \\ I will provide you an image and related information about the image…\\ Describe the image in as much detail as possible.
     \\ Image related information: $<$Image Tag$>$ 
     \\ Text information in the image: $<$OCR Text$>$ \textit{(if has)}
     \\ $<$Special Requirements$>$ \textit{(if has)} } \\
     \midrule 
    Numerical Calculation & \makecell[l]{Note that the image provides mathematical problems that may involve numerical values,\\ mathematical formulas, or graphics.} \\
    \midrule
    Brand Recognition & Try to identify the brand of the item in the image. \\
    \midrule 
    Posters & Try to identify which file/TV show the image comes from. \\
    \midrule 
    Landmark & Try to identify the landmark building or place in the image. \\ 
    \midrule 
    Meme Comprehension & Try to discern the intriguing aspects within the image. \\
    \midrule 
    Social Relation & Try to identify the relationship between the people in the image. \\
    \midrule 
    Spatial Relationship & \makecell[l]{
    Try to identify the spatial relationship between the objects in the image.} \\
    \bottomrule
    \end{tabular}
    \label{tab:caption-prompt} 
\end{table}

\subsection{Seed Question Collection} \label{sec:seed_ques}

Seed questions, serving as a reference for instruction generation in our data engine, directly influence the effectiveness of generated results. These seed questions should be generic, covering the majority of common instructions users may utilize. Furthermore, for different visual domains, corresponding seed questions should also have different focuses to clarify the context of the questions answered.
Our seed question templates can be divided into general questions and wildcard questions, Table \ref{table:seed-question-examples} illustrates some examples of our seed questions. Even for the same domain, the possible types of seed questions are diverse, including different asking methods and questions with or without wildcards.

\begin{table}[t!]
    \fontsize{8pt}{10pt}\selectfont
    \centering
    \caption{{Examples of seed questions in different domains.} \textbf{General} represents general questions; \textbf{Wildcard} represents questions containing placeholders.}
    \label{table:seed-question-examples}
    \begin{tabular}{c|l|p{10.7cm}}
    \toprule
    {Type} & {Domain} & {Seed Questions}  \\
    \midrule
    \multirow{3}{*}{\makecell[l]{General}} & species recognition & \makecell[l]{Identify the species in the image. 
    \\ What is the scientific name of this species?}
    \\
    \cmidrule{2-3}
     & image emotion & \makecell[l]{Which mood does this image convey?
     \\ Identify the emotion expressed in this image.} \\
    \cmidrule{2-3}
     & numerical calculation & \makecell[l]{Are the calculations in the image correct? 
     \\ Calculate the formulas in the picture.}
     \\
    
    \midrule 
    \multirow{3}{*}{Wildcard} & \makecell[l]{species recognition} & \makecell[l]{This is an $<$object$>$, which species does it belong to?
    \\ Is the scientific name of this $<$object$>$ $<$name$>$? }\\
    \cmidrule{2-3}
     & {image emotion }  & \makecell[l]{Does this image convey the emotion of $<$specific emotion$>$?
    \\ Is the emotion of $<$some object$>$ in the picture $<$positive/negative$>$?} \\
    \cmidrule{2-3}
     & numerical calculation & 
     \makecell[l]{What is the $<$area/volume$>$ of $<$the geometry$>$ in the image? 
     \\ What should the value of $<$variable$>$ in the picture be equal to? }\\
    \bottomrule
    \end{tabular}
\end{table}

When constructing \dataname, we aggregate a large amount of instruction data from existing open datasets and real users. Subsequently, experts summarize common questions in each domain based on a statistical analysis of the data to serve as seed questions. Overall, an average of about 10 seed questions are designed per domain. To ensure the effectiveness of the seed questions, a small batch of instructions is generated before the actual instruction generation process. This preliminary step can be used to verify the generated results and make appropriate modifications to the seed questions.

\subsection{Automatic Instruction Generation}\label{sec:inst_gen}

\begin{table}[t!]
    \fontsize{8pt}{10pt}\selectfont
    \centering
    \caption{{Key parts of prompts used for instruction generation in different domains.} \textbf{With Seed} represents the prompt used when generating instruction data based on seed questions; \textbf{No Seed} represents the prompt used when there is no generic question template; \textbf{Multi-Round} represents the prompt used to generate multi-round long visual question answering.}
    \label{table:prompt-examples}
    \renewcommand{\arraystretch}{1.5} 
    \begin{tabular}{l|p{1.8cm}|p{11.2cm}}
    \toprule
    {Type} & {Domains} & {Prompts}  \\
    \midrule
    With Seed & 
    \makecell[l]{numerical \\ calculation, 
    \\ attribute \\ recognition,
    \\ landmark,
    \\ \etc.}
    & \makecell[l]{Given a description of the image and a list of questions, you need to design 3 \\ $<$Question-type$>$ questions and corresponding answers related to the topic of \\ $<$Domain$>$…
     \\ Google OCR content: $<$OCR Result$>$
     \\ Image description: $<$Image Caption$>$
     \\ Question template: $<$Seed Questions$>$ 
     \\ You must output the generated questions, options, and answers in the following \\ format…}
    \\
    \midrule 
     \makecell[l]{No Seed} & 
    \makecell[l]{complex \\ reasoning,
    \\ commonsense \\ reasoning }
     & \makecell[l]{Given a description of the image, you need to ask 3 $<$Question-type$>$ questions \\ about the image that can be used in the $<$Domain$>$ task and generate \\ corresponding answers. 
    \\ You must output the generated questions, options, and answers in the following \\ format…}
    \\
    \midrule
    \makecell[l]{Multi-Round} & \makecell[l]{multi-round \\ long visual \\ question \\ answering} & \makecell[l]{Pretend that you have ``seen'' an image, based on the description provided below, \\ now you have two tasks:
    \\ Create 5 Questions using English: $<$Requests$>$ 
    \\ Answer the Questions using English: $<$Requests$>$ 
    \\ Example: $<$Example QA$>$ 
    \\ Image description: $<$Image Caption$>$…} \\
    \bottomrule
    \end{tabular}
\end{table}

After obtaining the seed question and image information, we utilize the text-only GPT-3.5 model to generate instruction data. We separately design generation pipelines for four types of questions: judgment, multiple-choice, and Short VQA and Long VQA. In the generation pipeline, for each image, its detailed caption and prior knowledge of the corresponding domain are used as input, and $N$ are randomly selected from the provided seed questions as references (with $N=3$ in our paper). Then guide GPT to generate instruction data according to the corresponding domain prompt. In order to enable the model to better distinguish the type of instructions, we add indicative utterances corresponding to the question type after each generated question. For example, ``Please choose the most appropriate option'' will be added to multiple-choice questions. The key part of prompts used is shown in Table \ref{table:prompt-examples} line 2.

It is worth noting that in some domains, seed questions may not be universally applicable to all images. For instance, in the numerical calculation domain, the seed questions for formula calculation and variable solving are distinctly different. To enhance the alignment between images and generated instructions, we categorize and match images with the corresponding seed questions. Moreover, the number of seed questions provided for reference is greater than the number of instructions that need to be generated, which can provide fault tolerance space for GPT, thus reducing the occurrence of unreasonable problems and hallucinations.

In commonsense reasoning and complex reasoning domains, there are diverse ways of questioning, hence we haven't collected seed questions. Instead, We employ a prompt for problems directly generated without a universal question template to instruct GPT in directly generating domain-relevant questions from detailed descriptions of the image.
The key part of prompts is shown in Table \ref{table:prompt-examples} line 3.

Due to factors such as language, culture, and individual habits, user instructions tend to be diverse. Therefore, we encourage the instruction questions generated by GPT to be of various styles, as long as they are semantically close to the seed question. 
To effectively mitigate hallucination during the generation process, we strictly enforce GPT to generate both questions and their answers at one time. The answers to questions must be correct and explicitly derived from the image information. In particular, for multiple-choice questions, GPT is also required to provide the four options corresponding to the question, ensuring that exactly one option is correct.

\subsection{Dataset Expansion}\label{sec:data_expan}

In order to further expand the diversity and versatility of \dataname, we also expand the dataset through other methods. On the one hand, we build a similar pipeline to generate multi-round long visual question answering (Multi-Round Long VQA) data to extend the instruction type. On the other hand, we screen and process some data from open-source datasets to extend the domain of our dataset.

\paragraph{Multi-Round Long VQA Data Expansion.} 
In practical user usage, multiple rounds of contextually related questions and answers are a common interaction mode. Multi-Round Long VQA data with rigorous logic and reasonable inference is crucial for model training. Such data aids in the learning of deeper semantic comprehension and inference capabilities, enabling models to perform more accurately and naturally in understanding questions and deducing answers. Therefore, we propose an automated pipeline for constructing Multi-Round Long VQA instructions, leveraging the powerful reasoning capabilities of GPT-4V. Similar to the instruction construction pipeline outlined in Section \ref{sec:inst_gen}, it also utilizes detailed image captions and prior knowledge as input. For the pipeline prompt, while strictly constraining GPT to adhere to the given information, we request it to generate 5 questions along with their corresponding correct answers each time. And these questions should have a continuous logical linkage and evolution between them. The key part of prompts used for Multi-Round Long VQA is shown in Table \ref{table:prompt-examples} line 4. It is worth noting that our multi-round response data is longer and more informative than other datasets. This will force the model to have a deeper understanding and rigorous analysis of the questions.

\paragraph{Other Source Data Expansion.} 
In order to further enrich the domain categories contained in our dataset and increase the diversity of instruction formats, we select some data from open-source datasets. This includes mathematics datasets~\cite{seo2015geos,chen2022unigeo,cao2022geoqa,lu2021geometry3k,lindstrom2022clevrmath,li2023superclevr, lu2022tabmwp}, charts and plots~\cite{kafle2018dvqa, kahou2017figureqa}, scientific figure~\cite{kembhavi2017tqa} and map chart~\cite{chang2022mapqa}. We then convert them into dialogue format and add them to \dataname.

\begin{figure}[t!]
    \centering
    \includegraphics[width=0.93\textwidth]{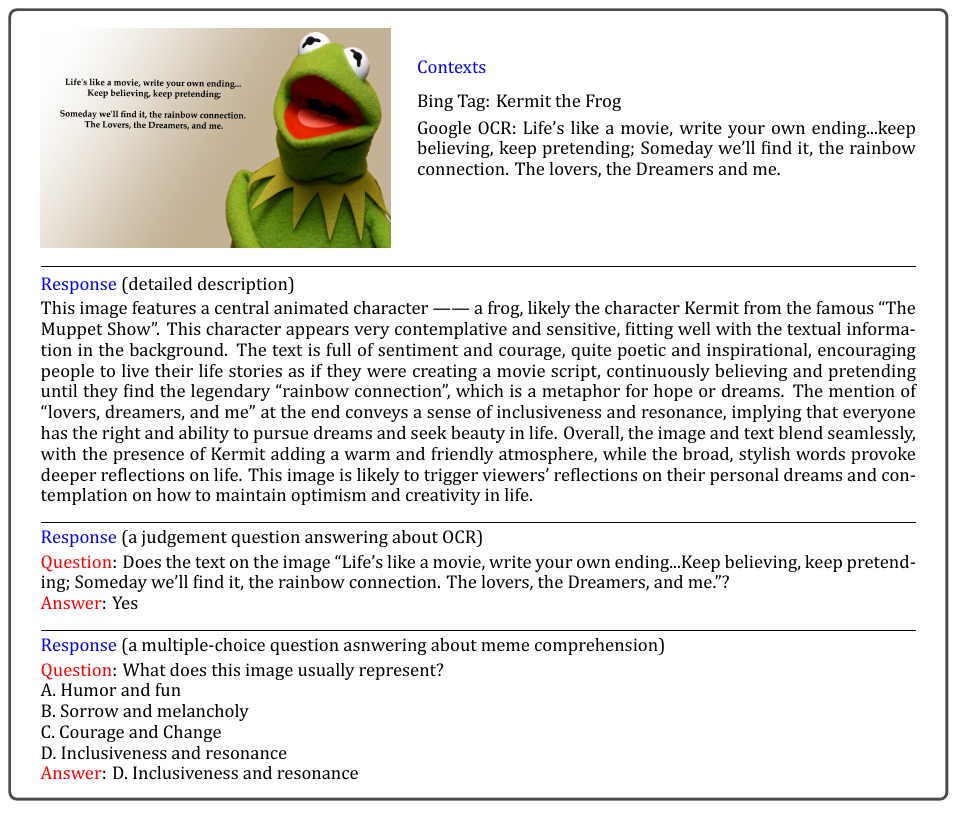}
    \caption{One example of the generated instruction data using our data engine. The top block shows the contexts such as the image tag from Bing and the OCR results obtained from Google, those contexts are used to prompt GPT to generate a detailed description of the given image. The second block is a detailed image description generated based on the image and context. The last two blocks show two different types of instruction data generated based on detailed image descriptions and seed questions.}
    \label{fig:example}
\end{figure}

\subsection{Manual Correction} 
The instruction data generated by our data engine has basically met the requirements of the instruction dataset. However, in bulk generation, some data inevitably contains problems such as hallucination, grammatical errors, or mismatches between instructions and domain. Therefore, additional manual corrections are necessary for the constructed data, with the cost significantly lower than that of manually constructing a complete dataset. Therefore, it is necessary to perform additional manual corrections on the constructed data, with costs much lower than manually building the dataset from scratch.

In this stage, we provide all data in the form of $<$\textit{image}, \textit{caption}, \textit{instruction answer}$>$ pair, as shown in Figure \ref{fig:example}, for manual correction by multiple professional annotation teams.
In order to ensure the quality of the dataset, we set acceptance criteria and hire annotation teams based on the characteristics of the instruction domain. 
For example, in the OCR domain, in addition to the regular annotation requirements, we additionally require the teams to pay attention to whether the text in the image is correctly recognized, whether the text content is comprehensively recognized, and whether the order of the text output corresponds to the position in the image, and so forth.
The teams inspect and modify each instruction data based on the image and caption to meet the strict acceptance criteria. In this process, the same batch of data is shuffled and undergoes three or more rounds of rework. Additionally, humans possess a greater understanding compared to GPT, resulting in manually corrected data exhibiting greater diversity and better alignment with human questioning styles.

\vspace{1em}

Within our data engine, the cost of processing and generating a detailed caption (averaging 200 words) for a 512px$\times$512px image using \textit{gpt-4-1106-vision-preview} is \$0.00885. On this basis, the average cost to generate an instruction using \textit{gpt-3.5-turbo-1106} is \$0.0004, with manual correction costing \$0.13. Therefore, for \dataname, handling 161K images and 973K instruction data requires approximately \$128, 304. In contrast, if we manually construct the dataset, each instruction costs around \$0.84, resulting in a total cost of approximately \$817, 320 for \dataname. Remarkably, leveraging our data engine costs only one-sixth of what it would cost to build it entirely by hand. This effectively demonstrates the cost-effectiveness of our data engine.

%% file: sections/4_experiments.tex
\section{Experiments}

\subsection{Experiment Setup}

To verify the effectiveness of our proposed dataset \dataname, we conduct a series of evaluation experiments. We follow the design of the advanced VLM architecture LLaVA-1.5~\cite{liu2023llava_1_5}, which mainly consists of three parts: the pre-trained vision encoder CLIP-ViT-L-336px~\cite{radford2021clip}, the pre-trained large language model Vicuna-v1.5~\cite{chiang2023vicuna}, and a 2 layer MLP projection.

\paragraph{Training Details.} We adopt the same two-stage training design as LLaVA-1.5. During the pre-training stage, we keep the vision encoder and large language model frozen and use the LCS-558K pre-training dataset to train the MLP projection. In the fine-tuning stage, we keep the vision encoder frozen and combine the LLaVA-665K instruction dataset with our \dataname to fine-tune the MLP projection and large language model.
Meanwhile, we use the same hyper-parameters as LLaVA-1.5.

\paragraph{Evaluation Benchmarks.}

We evaluate our visual instruction tuning dataset \dataname using the same benchmarks as LLaVA-1.5, including traditional academic visual
question answering benchmarks: 
\textbf{VQAv2}: VQA$^{\text{v2}}$~\cite{goyal2017vqav2},
\textbf{GQA}~\cite{hudson2019gqa},
\textbf{VizWiz}~\cite{gurari2018vizwiz},
\textbf{ScienceQA-Img}: SQA$^{\text{I}}$~\cite{lu2022sqa},
and \textbf{TextVQA}: VQA$^{\text{T}}$~\cite{singh2019textvqa}; 
comprehensive multi-modal evaluation benchmarks: 
\textbf{POPE}~\cite{li2023pope},
\textbf{MME}~\cite{fu2023mme},
\textbf{MMbench}: MMB~\cite{liu2023mmbench}, \textbf{MMbench-Chinese}: MMB$^{\text{CN}}$,
\textbf{SEED-Bench}: SEED~\cite{li2023seed}, 
\textbf{LLaVA-Bench} (In-the-Wild): LLaVA$^{\text{W}}$~\cite{liu2023llava} and 
\textbf{MM-Vet}~\cite{yu2023mmvet}.

\begin{table}[t!]
    \fontsize{8pt}{10pt}\selectfont
    \centering
    \caption{Comparison with state-of-the-art VLLMs on traditional VQA benchmarks. Priv: the data are private. $^{*}$ denotes the training images of the datasets are observed during training. The best results are marked in \textbf{bold}, and the second best results are \underline{underlined}.}
    \label{tab:traditional_vqa_bench}
    \begin{tabular}{p{3.2cm}>{\centering\arraybackslash}p{2.0cm}|ccccc}
            \toprule
            {Method} & {LLM} &  {VQA$^\text{v2}$} & {GQA} & {VizWiz} & {SQA$^\text{I}$} & {VQA$^\text{T}$} \\
            \midrule
            InstructBLIP~\cite{instructblip} & {Vicuna-7B}  & -- & 49.2 & 34.5 & 60.5 &  50.1  \\
            IDEFICS-9B~\cite{idefics2023} & {LLaMA-7B}  & 50.9 & 38.4 & 35.5 & – & 25.9 \\
            Qwen-VL~\cite{bai2023qwenvl} & {Qwen-7B}  & 78.8$^*$ & 59.3$^*$ & 35.2 & 67.1 & \textbf{63.8} \\
            Qwen-VL-chat~\cite{bai2023qwenvl} & {Qwen-7B}  & 78.2$^*$ & 57.5$^*$ & 38.9 & 68.2 & \underline{61.5} \\
            LLaVA-1.5~\cite{liu2023llava_1_5} & {Vicuna-7B}  & 78.5$^*$ & 62.0$^*$ & 50.0 & 66.8 & 58.2 \\
            \rowcolor{Gray}
            \makecell[l]{\fontsize{6pt}{10pt}\textbf{LLaVA-1.5}\\ \textbf{\fontsize{6pt}{10pt}+\dataname(ours)}} & {Vicuna-7B} & \underline{79.6}$^*$ & \underline{62.8}$^*$ & 53.3 &  71.2 & 57.7  \\
            \midrule
            BLIP-2~\cite{li2023blip2} & {Vicuna-13B}  & 65.0 & 41.0 & 19.6 & 61.0 & 42.5 \\
            InstructBLIP~\cite{instructblip} & {Vicuna-13B}  & -- & 49.5 & 33.4 & 63.1 & 50.7\\
            IDEFICS-80B~\cite{idefics2023} & {LLaMA-65B} & 60.0 & 45.2 & 36.0 & -- & 30.9  \\
            Shikra~\cite{chen2023shikra} & {Vicuna-13B}  & 77.4$^*$ & -- & -- & -- & -- \\
            LLaVA-1.5~\cite{liu2023llava_1_5} & {Vicuna-13B}  & \textbf{80.0}$^*$ & \textbf{63.3}$^*$ & \underline{53.6} & \underline{71.6} & 61.3 \\
            \rowcolor{Gray}
            \makecell[l]{\fontsize{6pt}{10pt}\textbf{LLaVA-1.5} \\ \textbf{\fontsize{6pt}{10pt}+\dataname(ours)}}
             & {Vicuna-13B}  & \textbf{80.0}$^*$ & 62.6$^*$ & \textbf{55.8} & \textbf{74.2} & 60.9 \\
            \bottomrule
        \end{tabular} 
\end{table}

\begin{table}[h!]
    \fontsize{8pt}{10pt}\selectfont
    \centering
    \caption{Comparison with state-of-the-art VLLMs on recent Multi-modal benchmarks.}
    \label{tab:lmm_bench}
    \begin{tabular}{p{3.2cm}>{\centering\arraybackslash}p{2.1cm}|ccccccc}
    \toprule
    {Method} & {LLM} &  {POPE} & {MME} & {MMB}  & {MMB$^\text{CN}$} & {SEED} & {LLaVA$^\text{W}$} & {MM-Vet} \\
    \midrule
    InstructBLIP~\cite{instructblip} & {Vicuna-7B}  & -- & -- & 36.0 & 23.7 & 53.4 & 60.9 & 26.2 \\
    IDEFICS-9B~\cite{idefics2023} & {LLaMA-7B}  & -- &  -- & 48.2  & 25.2 & -- & -- & -- \\
    Qwen-VL~\cite{bai2023qwenvl} & {Qwen-7B} & -- & -- & 38.2  & 7.4 & 56.3 & -- & -- \\
    Qwen-VL-chat~\cite{bai2023qwenvl} & {Qwen-7B}  & -- & 1487.5 & 60.6  &  56.7 & 58.2 & -- & -- \\
    LLaVA-1.5~\cite{liu2023llava_1_5} & {Vicuna-7B}  & 85.9 & 1510.7 & 64.3  & 58.3 & 58.6 & 63.4 & 30.5 \\ 
    \rowcolor{Gray}
    \makecell[l]{\fontsize{6pt}{10pt}\selectfont\textbf{LLaVA-1.5}\\ \fontsize{6pt}{10pt}\selectfont\textbf{+\dataname(ours)}} & {Vicuna-7B} & \underline{86.8} & \underline{1591.8} & \underline{69.8} & \underline{65.2} & \underline{64.6} & \underline{71.0} & 34.4 \\
    \midrule
    BLIP-2~\cite{li2023blip2} & {Vicuna-13B}  &  85.3 & 1293.8 & --  & -- & 46.4 & 38.1 & 22.4 \\
    InstructBLIP~\cite{instructblip} & {Vicuna-13B}  & 78.9 & 1212.8 & -- & -- & -- & 58.2 & 25.6 \\
    IDEFICS-80B~\cite{idefics2023} & {LLaMA-65B} & -- & -- & 54.5 & 38.1 & -- & -- & -- \\
    Shikra~\cite{chen2023shikra} & {Vicuna-13B}  & -- & -- & 58.8 & -- & -- & -- & -- \\
    LLaVA-1.5~\cite{liu2023llava_1_5} & {Vicuna-13B}  & 85.9 & 1531.3 & 67.7 & 63.6 & 61.6 & 70.7 & \underline{35.4} \\
    \rowcolor{Gray}
    \makecell[l]{\fontsize{6pt}{10pt}\selectfont \textbf{LLaVA-1.5}\\ 
 \fontsize{6pt}{10pt}\selectfont \textbf{+\dataname(ours)}} & {Vicuna-13B}  & \textbf{86.9} & \textbf{1626.2} & \textbf{72.1} & \textbf{68.0}  & \textbf{64.7} & \textbf{74.5} & \textbf{37.9} \\
    \bottomrule
    \end{tabular}
\end{table}

\subsection{Main Results}

As shown in Table \ref{tab:traditional_vqa_bench} and Table \ref{tab:lmm_bench}, in quantitative comparisons with leading VLLMs, our 7B and 13B models significantly outperform LLaVA-1.5 models across various benchmarks, including both academic visual question answering and multi-modal evaluation benchmarks.
It is worth noting that our 13B model achieves state-of-the-art performance on 10 out of 12 benchmarks.

\paragraph{Results of Visual Question Answering Benchmarks.} 
On general VQA benchmarks, Our 13B model has shown significant improvements over LLaVA-1.5 models, particularly in VizWiz and ScienceQA, especially ScienceQA exhibiting an improvement of nearly 3\% compared to LLaVA-1.5. Additionally, our model has demonstrated competitive performance in VQAv2, GQA, and TextVQA benchmarks as well.

\paragraph{Results of Multi-modal Benchmarks.} In recent comprehensive multi-modal benchmarks, which contain fine-grained multi-modal tasks across a wide range of tasks. Our model achieves state-of-the-art performance on these benchmarks, surpassing LLaVA-1.5 comprehensively. Specifically, we achieve a substantial gain of 94.9 points (1626.2 vs. 1531.3) on MME and an impressive improvement of 4.4 points (68.0 vs. 63.6) on MMBench-CN. Furthermore, our model exhibits significant enhancements over LLaVA-1.5 on other multi-modal benchmarks, including POPE, MMBench, SEED, LLaVA-Bench (In-the-Wild), and MM-Vet. These results highlight the effectiveness of \dataname.

\vspace{1.5em}

We attribute these performance improvements to the advantages of \dataname, including image diversity, instruction diversity, and high-quality annotations. The dataset includes images from a broader range of domains, enabling the model to have greater generalization capability. Simultaneously, detailed and precise image captions and strictly set prompts when generating instructions effectively reduce model illusion and deviation questions. Moreover, our data engine ensures that the generated instructions encompass the four common question types and feature complex and varied sentence structures. This compels the model to deeply understand the essence of tasks rather than merely learning surface-level sentence patterns. Consequently, it exhibits satisfactory responses to different instruction formats across VQA and multi-modal benchmarks.

\subsection{Ablation Studies}

While keeping pre-training data the same, we also report the quantitative results of varying the fine-tuning data. This can provide insights for more efficient use of \dataname later.

\begin{figure}[t!]
    \centering
    \includegraphics[width=\textwidth]{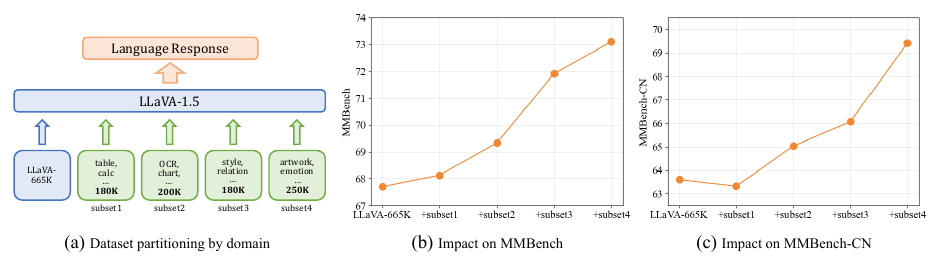}
    \caption{Performance on MMBench/MMbench-CN versus the number of training domains.}
    \label{fig:ablation-domain}
\end{figure}

\paragraph{Effect of Domain Diversity in \dataname.} To investigate the impact of domain diversity on model performance, we conduct ablation experiments on the number of domains. We randomly divide the data of \dataname into subsets with similar data size by domain. Subsequently, we employ the LLaVA-1.5 13B model pre-trained with the same settings. Only during fine-tuning stages do we incrementally add new data subsets for each experiment. Figure \ref{fig:ablation-domain} shows the relationship between the performance of the two comprehensive evaluation benchmarks, MMBench and MMBench-CN, and the number of \dataname fields used in the instruction fine-tuning stage. It's evident that increasing the number of domains can significantly improve the performance of the model in both Chinese and English. Furthermore, this performance improvement shows a linear increasing trend overall. These findings robustly confirm the rationality of our domain categorization and validate the effectiveness of domain diversity.

\begin{figure}[t!]
    \centering
    \includegraphics[width=0.85\textwidth]{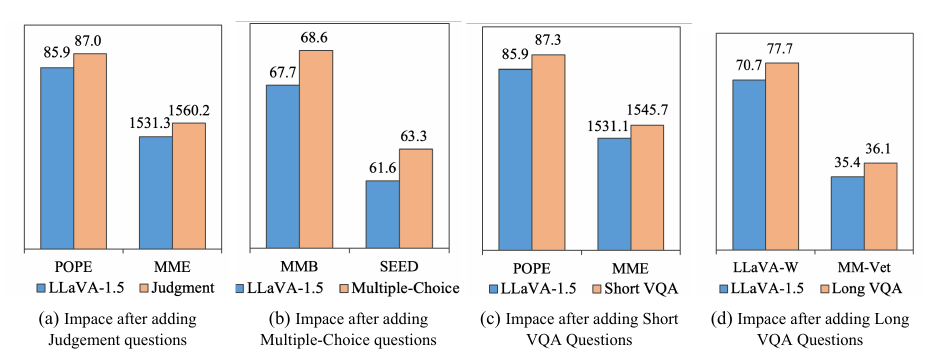}
    \caption{Performance on multi-modal benchmarks versus the question types.}
    \label{fig:ablation-question-type}
\end{figure}

\paragraph{Effect of Question Types Diversity in \dataname.} We analyze the impact of different question types used in the instruction fine-tuning stage on model performance. The model used in the experiment is LLaVA-1.5 13B, which is pre-trained without any instruction fine-tuning. We categorize the data in \dataname based on question types into judgment, multiple-choice, Long VQA, and Short VQA. In each experiment, we randomly select 150K samples from only one type of question data and combine them with the LLaVA-665K instruction dataset to fine-tune the model. In comparison to the baseline utilizing solely the LLaVA-665K dataset, the four sub-figures depicted in Figure \ref{fig:ablation-question-type} showcase two prominent benchmarks for distinct types of questions each. Specifically, the inclusion of Long VQA data leads to substantial improvements of 7.0 and 0.7 points in the LLaVA-Bench (In-the-Wild) and MM-Vet benchmarks, respectively. The utilization of multi-choice data significantly enhances the MMBench and SEED benchmarks by 0.9 and 1.7 points, respectively. This strongly indicates that the diversity of question types is crucial for models to effectively comprehend tasks.

\begin{wraptable}{h}{0.45\textwidth}
   \fontsize{8pt}{10pt}\selectfont
    \centering
    \caption{{Performance on LLaVA-Wild and MM-Vet versus Multi-Round Long VQA.} \textbf{NoMR} represents a model tuned without Multi-Round Long VQA.}
    \label{tab:multi-round-diaglogue-1}
    \begin{tabular}{l|cc}
    \toprule
    {Method}  & {LLaVA$^\text{W}$} & {MM-Vet} \\
    \midrule
    LLaVA-1.5 & 70.7 & 35.4 \\
    \midrule
    \makecell[l]{\fontsize{6pt}{10pt}\selectfont{LLaVA-1.5}\\ \fontsize{6pt}{10pt}\selectfont{+\dataname-NoMR (ours)}}  & 72.3 & 37.4 \\
    \midrule
    \makecell[l]{\fontsize{6pt}{10pt}\selectfont{LLaVA-1.5}\\ \fontsize{6pt}{10pt}\selectfont{+\dataname(ours)}} & \textbf{74.5} & \textbf{37.9} \\
    \bottomrule
    \end{tabular}
    \vspace{-0.2cm}
\end{wraptable}

\paragraph{Effect of Multi-Round Long VQA in \dataname.} 
As shown in Table \ref{tab:multi-round-diaglogue-1}, we attempt to remove the Multi-Round Long VQA data from \dataname to assess its impact on the model. For a fair comparison, we train the model using the same settings and images, with the only difference being that the Multi-Round Long VQA instructions are removed from \dataname for all images. The results show that Multi-Round Long VQA data can make the model achieve significant gains, with improvements of 2.2 and 0.5 points on the LLaVA-Bench (In-the-Wild) and MM-Vet benchmarks, respectively. We speculate that this is because the Multi-Round Long VQA data effectively trains the model to handle long contexts, while the generation of long outputs also effectively enhances the model's inference capabilities for complex tasks. This proves the necessity of adding high-quality Multi-Round Long VQA data to the instruction fine-tuning dataset.

\subsection{Visualizations} 

\begin{figure}[h!]
    \centering
    \includegraphics[width=0.9\textwidth]{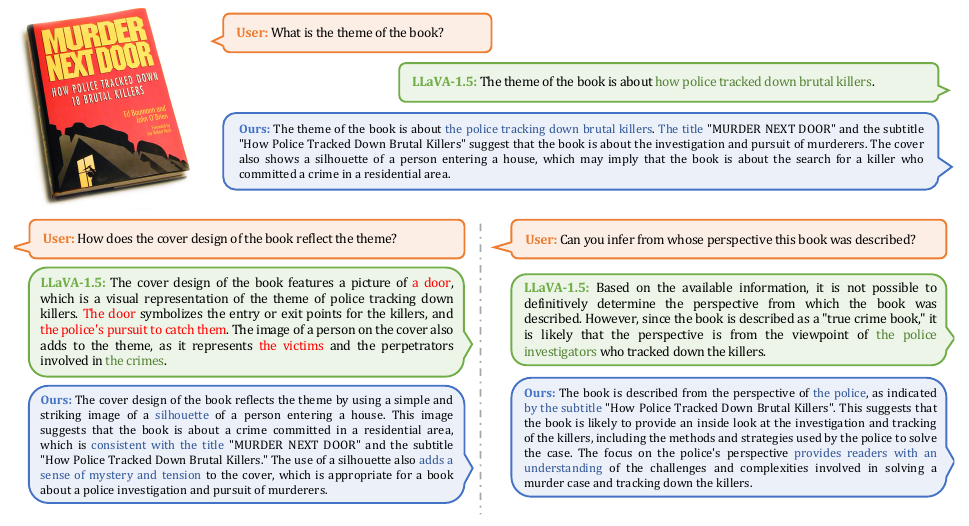}
    \caption{Visualization of outputs comparison between LLaVA-1.5 and our model. }
    \label{fig:visualization}
\end{figure}

We visualize some example outputs of LLaVA-1.5 and our model in Figure \ref{fig:visualization}. Note that the text marked with the corresponding background color represents the correct reasoning process and results, while the text marked in red indicates hallucinations. We can see that with additional instruction fine-tuning on \dataname, our model is better able to reason with contextual information in Multi-Round Long VQA, and its answers contain a more explicit reasoning process. In particular, our model is able to more accurately recognize text in pictures and understand the spatial relationships of image objects, which is significantly beneficial for logical reasoning.

%% file: sections/5_conclusion.tex
\section{Conclusion}

In this paper, we constructed a high-quality, diverse visual instruction tuning dataset \dataname, which consists of 973K instructions from 24 domains. Specifically, \dataname contains diverse question forms and types, including Judgment, Multiple-Choice, Long VQA, and Short VQA.
To construct \dataname, we propose an instruction generation data engine that leverages GPT-4V, GPT-3.5, and manual correction. Our data engine enables semi-automatic, low-cost, and multi-domain instruction generation. Compared to manual construction, our data engine's cost is only 1/6 of manual annotation while ensuring annotation quality and data diversity.
Then, we incorporate \dataname into the instruction fine-tuning phase of LLaVA-1.5 to evaluate its effectiveness. The results show that our model demonstrates impressive performance across multiple multi-modal benchmarks. 
Additionally, we also perform comprehensive ablation experiments to analysis the impacts of varying the fine-tuning data on VLLMs. These results clearly demonstrate that \dataname benefits visual instruction tuning.

%% file: main.bbl
\begin{thebibliography}{10}

\bibitem{alayrac2022flamingo}
Jean-Baptiste Alayrac, Jeff Donahue, Pauline Luc, Antoine Miech, Iain Barr, Yana Hasson, Karel Lenc, Arthur Mensch, Katherine Millican, Malcolm Reynolds, et~al.
\newblock Flamingo: a visual language model for few-shot learning.
\newblock {\em NeurIPS}, 2022.

\bibitem{bai2023qwenvl}
Jinze Bai, Shuai Bai, Shusheng Yang, Shijie Wang, Sinan Tan, Peng Wang, Junyang Lin, Chang Zhou, and Jingren Zhou.
\newblock Qwen-vl: A frontier large vision-language model with versatile abilities.
\newblock {\em arXiv preprint arXiv:2308.12966}, 2023.

\bibitem{vlmo}
Hangbo Bao, Wenhui Wang, Li~Dong, Qiang Liu, Owais~Khan Mohammed, Kriti Aggarwal, Subhojit Som, Songhao Piao, and Furu Wei.
\newblock {VLMo}: Unified vision-language pre-training with mixture-of-modality-experts.
\newblock In {\em Advances in Neural Information Processing Systems}, 2022.

\bibitem{bao2022vlbeit}
Hangbo Bao, Wenhui Wang, Li~Dong, and Furu Wei.
\newblock Vl-beit: Generative vision-language pretraining, 2022.

\bibitem{cao2022geoqa}
Jie Cao and Jing Xiao.
\newblock An augmented benchmark dataset for geometric question answering through dual parallel text encoding.
\newblock In {\em Proceedings of the 29th International Conference on Computational Linguistics}, pages 1511--1520, 2022.

\bibitem{chang2022mapqa}
Shuaichen Chang, David Palzer, Jialin Li, Eric Fosler-Lussier, and Ningchuan Xiao.
\newblock Mapqa: A dataset for question answering on choropleth maps.
\newblock {\em arXiv preprint arXiv:2211.08545}, 2022.

\bibitem{chen2022unigeo}
Jiaqi Chen, Tong Li, Jinghui Qin, Pan Lu, Liang Lin, Chongyu Chen, and Xiaodan Liang.
\newblock Unigeo: Unifying geometry logical reasoning via reformulating mathematical expression.
\newblock {\em arXiv preprint arXiv:2212.02746}, 2022.

\bibitem{chen2022visualgpt}
Jun Chen, Han Guo, Kai Yi, Boyang Li, and Mohamed Elhoseiny.
\newblock Visualgpt: Data-efficient adaptation of pretrained language models for image captioning.
\newblock In {\em Proceedings of the IEEE/CVF Conference on Computer Vision and Pattern Recognition}, pages 18030--18040, 2022.

\bibitem{chen2023shikra}
Keqin Chen, Zhao Zhang, Weili Zeng, Richong Zhang, Feng Zhu, and Rui Zhao.
\newblock Shikra: Unleashing multimodal llm's referential dialogue magic.
\newblock {\em arXiv preprint arXiv:2306.15195}, 2023.

\bibitem{chen2023sharegpt4v}
Lin Chen, Jisong Li, Xiaoyi Dong, Pan Zhang, Conghui He, Jiaqi Wang, Feng Zhao, and Dahua Lin.
\newblock Sharegpt4v: Improving large multi-modal models with better captions.
\newblock {\em arXiv preprint arXiv:2311.12793}, 2023.

\bibitem{chen2022pali}
Xi~Chen, Xiao Wang, Soravit Changpinyo, AJ~Piergiovanni, Piotr Padlewski, Daniel Salz, Sebastian Goodman, Adam Grycner, Basil Mustafa, Lucas Beyer, et~al.
\newblock Pali: A jointly-scaled multilingual language-image model.
\newblock In {\em ICLR}, 2022.

\bibitem{chen2024internvl_v1_5}
Zhe Chen, Weiyun Wang, Hao Tian, Shenglong Ye, Zhangwei Gao, Erfei Cui, Wenwen Tong, Kongzhi Hu, Jiapeng Luo, Zheng Ma, et~al.
\newblock How far are we to gpt-4v? closing the gap to commercial multimodal models with open-source suites.
\newblock {\em arXiv preprint arXiv:2404.16821}, 2024.

\bibitem{chen2023internvl}
Zhe Chen, Jiannan Wu, Wenhai Wang, Weijie Su, Guo Chen, Sen Xing, Muyan Zhong, Qinglong Zhang, Xizhou Zhu, Lewei Lu, Bin Li, Ping Luo, Tong Lu, Yu~Qiao, and Jifeng Dai.
\newblock Internvl: Scaling up vision foundation models and aligning for generic visual-linguistic tasks.
\newblock {\em arXiv preprint arXiv:2312.14238}, 2023.

\bibitem{chiang2023vicuna}
Wei-Lin Chiang, Zhuohan Li, Zi~Lin, Ying Sheng, Zhanghao Wu, Hao Zhang, Lianmin Zheng, Siyuan Zhuang, Yonghao Zhuang, Joseph~E Gonzalez, et~al.
\newblock Vicuna: An open-source chatbot impressing gpt-4 with 90\%* chatgpt quality.
\newblock {\em See https://vicuna. lmsys. org (accessed 14 April 2023)}, 2(3):6, 2023.

\bibitem{instructblip}
Wenliang Dai, Junnan Li, Dongxu Li, AnthonyMeng Huat, Junqi Zhao, Weisheng Wang, Boyang Li, Pascale Fung, and Steven Hoi.
\newblock Instructblip: Towards general-purpose vision-language models with instruction tuning.
\newblock {\em arXiv preprint arXiv:2305.06500}, 2023.

\bibitem{dong2024ixc4khd}
Xiaoyi Dong, Pan Zhang, Yuhang Zang, Yuhang Cao, Bin Wang, Linke Ouyang, Songyang Zhang, Haodong Duan, Wenwei Zhang, Yining Li, et~al.
\newblock Internlm-xcomposer2-4khd: A pioneering large vision-language model handling resolutions from 336 pixels to 4k hd.
\newblock {\em arXiv preprint arXiv:2404.06512}, 2024.

\bibitem{fang2022eva}
Yuxin Fang, Wen Wang, Binhui Xie, Quan Sun, Ledell Wu, Xinggang Wang, Tiejun Huang, Xinlong Wang, and Yue Cao.
\newblock Eva: Exploring the limits of masked visual representation learning at scale.
\newblock {\em arXiv preprint arXiv:2211.07636}, 2022.

\bibitem{fu2023mme}
Chaoyou Fu, Peixian Chen, Yunhang Shen, Yulei Qin, Mengdan Zhang, Xu~Lin, Zhenyu Qiu, Wei Lin, Jinrui Yang, Xiawu Zheng, et~al.
\newblock Mme: A comprehensive evaluation benchmark for multimodal large language models.
\newblock {\em arXiv preprint arXiv:2306.13394}, 2023.

\bibitem{goyal2017vqav2}
Yash Goyal, Tejas Khot, Douglas Summers-Stay, Dhruv Batra, and Devi Parikh.
\newblock Making the v in vqa matter: Elevating the role of image understanding in visual question answering.
\newblock In {\em CVPR}, 2017.

\bibitem{gurari2018vizwiz}
Danna Gurari, Qing Li, Abigale~J Stangl, Anhong Guo, Chi Lin, Kristen Grauman, Jiebo Luo, and Jeffrey~P Bigham.
\newblock Vizwiz grand challenge: Answering visual questions from blind people.
\newblock In {\em CVPR}, 2018.

\bibitem{hudson2019gqa}
Drew~A Hudson and Christopher~D Manning.
\newblock Gqa: A new dataset for real-world visual reasoning and compositional question answering.
\newblock In {\em CVPR}, 2019.

\bibitem{idefics2023}
{IDEFICS}.
\newblock Introducing idefics: An open reproduction of state-of-the-art visual language model.
\newblock \url{https://huggingface.co/blog/idefics}, 2023.

\bibitem{jia2021align}
Chao Jia, Yinfei Yang, Ye~Xia, Yi-Ting Chen, Zarana Parekh, Hieu Pham, Quoc Le, Yun-Hsuan Sung, Zhen Li, and Tom Duerig.
\newblock Scaling up visual and vision-language representation learning with noisy text supervision.
\newblock In {\em Int. Conf. Mach. Learn.}, 2021.

\bibitem{kafle2018dvqa}
Kushal Kafle, Brian Price, Scott Cohen, and Christopher Kanan.
\newblock Dvqa: Understanding data visualizations via question answering.
\newblock In {\em Proceedings of the IEEE conference on computer vision and pattern recognition}, pages 5648--5656, 2018.

\bibitem{kahou2017figureqa}
Samira~Ebrahimi Kahou, Vincent Michalski, Adam Atkinson, {\'A}kos K{\'a}d{\'a}r, Adam Trischler, and Yoshua Bengio.
\newblock Figureqa: An annotated figure dataset for visual reasoning.
\newblock {\em arXiv preprint arXiv:1710.07300}, 2017.

\bibitem{kembhavi2017tqa}
Aniruddha Kembhavi, Minjoon Seo, Dustin Schwenk, Jonghyun Choi, Ali Farhadi, and Hannaneh Hajishirzi.
\newblock Are you smarter than a sixth grader? textbook question answering for multimodal machine comprehension.
\newblock In {\em Proceedings of the IEEE Conference on Computer Vision and Pattern recognition}, pages 4999--5007, 2017.

\bibitem{lai2023lisa}
Xin Lai, Zhuotao Tian, Yukang Chen, Yanwei Li, Yuhui Yuan, Shu Liu, and Jiaya Jia.
\newblock Lisa: Reasoning segmentation via large language model.
\newblock {\em arXiv preprint arXiv:2308.00692}, 2023.

\bibitem{li2023mimic}
Bo~Li, Yuanhan Zhang, Liangyu Chen, Jinghao Wang, Fanyi Pu, Jingkang Yang, Chunyuan Li, and Ziwei Liu.
\newblock Mimic-it: Multi-modal in-context instruction tuning.
\newblock {\em arXiv preprint arXiv:2306.05425}, 2023.

\bibitem{li2023seed}
Bohao Li, Rui Wang, Guangzhi Wang, Yuying Ge, Yixiao Ge, and Ying Shan.
\newblock Seed-bench: Benchmarking multimodal llms with generative comprehension.
\newblock {\em arXiv preprint arXiv:2307.16125}, 2023.

\bibitem{li2023uni}
Hao Li, Jinguo Zhu, Xiaohu Jiang, Xizhou Zhu, Hongsheng Li, Chun Yuan, Xiaohua Wang, Yu~Qiao, Xiaogang Wang, Wenhai Wang, et~al.
\newblock Uni-perceiver v2: A generalist model for large-scale vision and vision-language tasks.
\newblock In {\em Proceedings of the IEEE/CVF Conference on Computer Vision and Pattern Recognition}, pages 2691--2700, 2023.

\bibitem{li2023blip2}
Junnan Li, Dongxu Li, Silvio Savarese, and Steven Hoi.
\newblock Blip-2: Bootstrapping language-image pre-training with frozen image encoders and large language models.
\newblock {\em arXiv preprint arXiv:2301.12597}, 2023.

\bibitem{li2021albef}
Junnan Li, Ramprasaath Selvaraju, Akhilesh Gotmare, Shafiq Joty, Caiming Xiong, and Steven Chu~Hong Hoi.
\newblock Align before fuse: Vision and language representation learning with momentum distillation.
\newblock {\em NeurIPS}, 2021.

\bibitem{2023videochat}
Kunchang Li, Yinan He, Yi~Wang, Yizhuo Li, Wenhai Wang, Ping Luo, Yali Wang, Limin Wang, and Yu~Qiao.
\newblock Videochat: Chat-centric video understanding.
\newblock {\em arXiv preprint arXiv:2305.06355}, 2023.

\bibitem{li2023m}
Lei Li, Yuwei Yin, Shicheng Li, Liang Chen, Peiyi Wang, Shuhuai Ren, Mukai Li, Yazheng Yang, Jingjing Xu, Xu~Sun, et~al.
\newblock M $^{3}$ it: A large-scale dataset towards multi-modal multilingual instruction tuning.
\newblock {\em arXiv preprint arXiv:2306.04387}, 2023.

\bibitem{li2023pope}
Yifan Li, Yifan Du, Kun Zhou, Jinpeng Wang, Wayne~Xin Zhao, and Ji-Rong Wen.
\newblock Evaluating object hallucination in large vision-language models.
\newblock {\em EMNLP}, 2023.

\bibitem{li2023superclevr}
Zhuowan Li, Xingrui Wang, Elias Stengel-Eskin, Adam Kortylewski, Wufei Ma, Benjamin Van~Durme, and Alan~L Yuille.
\newblock Super-clevr: A virtual benchmark to diagnose domain robustness in visual reasoning.
\newblock In {\em Proceedings of the IEEE/CVF Conference on Computer Vision and Pattern Recognition}, pages 14963--14973, 2023.

\bibitem{lin2014microsoft}
Tsung-Yi Lin, Michael Maire, Serge Belongie, James Hays, Pietro Perona, Deva Ramanan, Piotr Doll{\'a}r, and C~Lawrence Zitnick.
\newblock Microsoft coco: Common objects in context.
\newblock In {\em ECCV}, 2014.

\bibitem{lindstrom2022clevrmath}
Adam~Dahlgren Lindstr{\"o}m and Savitha~Sam Abraham.
\newblock Clevr-math: A dataset for compositional language, visual and mathematical reasoning.
\newblock {\em arXiv preprint arXiv:2208.05358}, 2022.

\bibitem{liu2023lrv}
Fuxiao Liu, Kevin Lin, Linjie Li, Jianfeng Wang, Yaser Yacoob, and Lijuan Wang.
\newblock Aligning large multi-modal model with robust instruction tuning.
\newblock {\em arXiv preprint arXiv:2306.14565}, 2023.

\bibitem{liu2023llava_1_5}
Haotian Liu, Chunyuan Li, Yuheng Li, and Yong~Jae Lee.
\newblock Improved baselines with visual instruction tuning.
\newblock {\em arXiv preprint arXiv:2310.03744}, 2023.

\bibitem{liu2023llava}
Haotian Liu, Chunyuan Li, Qingyang Wu, and Yong~Jae Lee.
\newblock Visual instruction tuning.
\newblock {\em arXiv preprint arXiv:2304.08485}, 2023.

\bibitem{liu2023mmbench}
Yuan Liu, Haodong Duan, Yuanhan Zhang, Bo~Li, Songyang Zhang, Wangbo Zhao, Yike Yuan, Jiaqi Wang, Conghui He, Ziwei Liu, et~al.
\newblock Mmbench: Is your multi-modal model an all-around player?
\newblock {\em arXiv preprint arXiv:2307.06281}, 2023.

\bibitem{liu2023interngpt}
Zhaoyang Liu, Yinan He, Wenhai Wang, Weiyun Wang, Yi~Wang, Shoufa Chen, Qinglong Zhang, Yang Yang, Qingyun Li, Jiashuo Yu, et~al.
\newblock Interngpt: Solving vision-centric tasks by interacting with chatbots beyond language.
\newblock {\em arXiv preprint arXiv:2305.05662}, 2023.

\bibitem{lu2021geometry3k}
Pan Lu, Ran Gong, Shibiao Jiang, Liang Qiu, Siyuan Huang, Xiaodan Liang, and Song-Chun Zhu.
\newblock Inter-gps: Interpretable geometry problem solving with formal language and symbolic reasoning.
\newblock {\em arXiv preprint arXiv:2105.04165}, 2021.

\bibitem{lu2022sqa}
Pan Lu, Swaroop Mishra, Tanglin Xia, Liang Qiu, Kai-Wei Chang, Song-Chun Zhu, Oyvind Tafjord, Peter Clark, and Ashwin Kalyan.
\newblock Learn to explain: Multimodal reasoning via thought chains for science question answering.
\newblock {\em NeurIPS}, 2022.

\bibitem{lu2022tabmwp}
Pan Lu, Liang Qiu, Kai-Wei Chang, Ying~Nian Wu, Song-Chun Zhu, Tanmay Rajpurohit, Peter Clark, and Ashwin Kalyan.
\newblock Dynamic prompt learning via policy gradient for semi-structured mathematical reasoning.
\newblock {\em arXiv preprint arXiv:2209.14610}, 2022.

\bibitem{luo2023valley}
Ruipu Luo, Ziwang Zhao, Min Yang, Junwei Dong, Minghui Qiu, Pengcheng Lu, Tao Wang, and Zhongyu Wei.
\newblock Valley: Video assistant with large language model enhanced ability, 2023.

\bibitem{lyu2023macaw}
Chenyang Lyu, Minghao Wu, Longyue Wang, Xinting Huang, Bingshuai Liu, Zefeng Du, Shuming Shi, and Zhaopeng Tu.
\newblock Macaw-llm: Multi-modal language modeling with image, audio, video, and text integration.
\newblock {\em arXiv preprint arXiv:2306.09093}, 2023.

\bibitem{chatgpt}
OpenAI.
\newblock Chatgpt, 2022.

\bibitem{openai2023gpt4}
OpenAI.
\newblock Gpt-4 technical report.
\newblock {\em arXiv preprint arXiv:2303.08774}, 2023.

\bibitem{gpt4v}
OpenAI.
\newblock Gpt-4v(ision) system card.
\newblock 2023.

\bibitem{openai2022chatgpt}
TB~OpenAI.
\newblock Chatgpt: Optimizing language models for dialogue.
\newblock {\em OpenAI}, 2022.

\bibitem{ouyang2022training}
Long Ouyang, Jeffrey Wu, Xu~Jiang, Diogo Almeida, Carroll Wainwright, Pamela Mishkin, Chong Zhang, Sandhini Agarwal, Katarina Slama, Alex Ray, et~al.
\newblock Training language models to follow instructions with human feedback.
\newblock {\em Advances in neural information processing systems}, 35:27730--27744, 2022.

\bibitem{radford2021clip}
Alec Radford, Jong~Wook Kim, Chris Hallacy, Aditya Ramesh, Gabriel Goh, Sandhini Agarwal, Girish Sastry, Amanda Askell, Pamela Mishkin, Jack Clark, et~al.
\newblock Learning transferable visual models from natural language supervision.
\newblock In {\em ICML}, 2021.

\bibitem{reid2024gemini_1_5}
Machel Reid, Nikolay Savinov, Denis Teplyashin, Dmitry Lepikhin, Timothy Lillicrap, Jean-baptiste Alayrac, Radu Soricut, Angeliki Lazaridou, Orhan Firat, Julian Schrittwieser, et~al.
\newblock Gemini 1.5: Unlocking multimodal understanding across millions of tokens of context.
\newblock {\em arXiv preprint arXiv:2403.05530}, 2024.

\bibitem{Ren2023TimeChat}
Shuhuai Ren, Linli Yao, Shicheng Li, Xu~Sun, and Lu~Hou.
\newblock Timechat: A time-sensitive multimodal large language model for long video understanding.
\newblock {\em ArXiv}, abs/2312.02051, 2023.

\bibitem{schuhmann2022laion5b}
Christoph Schuhmann, Romain Beaumont, Richard Vencu, Cade Gordon, Ross Wightman, Mehdi Cherti, Theo Coombes, Aarush Katta, Clayton Mullis, Mitchell Wortsman, et~al.
\newblock Laion-5b: An open large-scale dataset for training next generation image-text models.
\newblock {\em NeurIPS}, 2022.

\bibitem{seo2015geos}
Minjoon Seo, Hannaneh Hajishirzi, Ali Farhadi, Oren Etzioni, and Clint Malcolm.
\newblock Solving geometry problems: Combining text and diagram interpretation.
\newblock In {\em Proceedings of the 2015 conference on empirical methods in natural language processing}, pages 1466--1476, 2015.

\bibitem{singh2019textvqa}
Amanpreet Singh, Vivek Natarajan, Meet Shah, Yu~Jiang, Xinlei Chen, Dhruv Batra, Devi Parikh, and Marcus Rohrbach.
\newblock Towards vqa models that can read.
\newblock In {\em CVPR}, 2019.

\bibitem{su2019vl_bert}
Weijie Su, Xizhou Zhu, Yue Cao, Bin Li, Lewei Lu, Furu Wei, and Jifeng Dai.
\newblock Vl-bert: Pre-training of generic visual-linguistic representations.
\newblock {\em arXiv preprint arXiv:1908.08530}, 2019.

\bibitem{InstructDoc2024}
Ryota Tanaka, Taichi Iki, Kyosuke Nishida, Kuniko Saito, and Jun Suzuki.
\newblock Instructdoc: A dataset for zero-shot generalization of visual document understanding with instructions.
\newblock In {\em AAAI}, 2024.

\bibitem{team2023gemini}
Gemini Team, Rohan Anil, Sebastian Borgeaud, Yonghui Wu, Jean-Baptiste Alayrac, Jiahui Yu, Radu Soricut, Johan Schalkwyk, Andrew~M Dai, Anja Hauth, et~al.
\newblock Gemini: a family of highly capable multimodal models.
\newblock {\em arXiv preprint arXiv:2312.11805}, 2023.

\bibitem{2023internlm}
InternLM Team.
\newblock Internlm: A multilingual language model with progressively enhanced capabilities.
\newblock \url{https://github.com/InternLM/InternLM}, 2023.

\bibitem{tian2024mminterleaved}
Changyao Tian, Xizhou Zhu, Yuwen Xiong, Weiyun Wang, Zhe Chen, Wenhai Wang, Yuntao Chen, Lewei Lu, Tong Lu, Jie Zhou, et~al.
\newblock Mm-interleaved: Interleaved image-text generative modeling via multi-modal feature synchronizer.
\newblock {\em arXiv preprint arXiv:2401.10208}, 2024.

\bibitem{touvron2023llama}
Hugo Touvron, Thibaut Lavril, Gautier Izacard, Xavier Martinet, Marie-Anne Lachaux, Timoth{\'e}e Lacroix, Baptiste Rozi{\`e}re, Naman Goyal, Eric Hambro, Faisal Azhar, et~al.
\newblock Llama: Open and efficient foundation language models.
\newblock {\em arXiv preprint arXiv:2302.13971}, 2023.

\bibitem{touvron2023llama2}
Hugo Touvron, Louis Martin, Kevin Stone, Peter Albert, Amjad Almahairi, Yasmine Babaei, Nikolay Bashlykov, Soumya Batra, Prajjwal Bhargava, Shruti Bhosale, et~al.
\newblock Llama 2: Open foundation and fine-tuned chat models.
\newblock {\em arXiv preprint arXiv:2307.09288}, 2023.

\bibitem{tsimpoukelli2021multimodal}
Maria Tsimpoukelli, Jacob~L Menick, Serkan Cabi, SM~Eslami, Oriol Vinyals, and Felix Hill.
\newblock Multimodal few-shot learning with frozen language models.
\newblock {\em Advances in Neural Information Processing Systems}, 34:200--212, 2021.

\bibitem{wang2024allseeing_v2}
Weiyun Wang, Yiming Ren, Haowen Luo, Tiantong Li, Chenxiang Yan, Zhe Chen, Wenhai Wang, Qingyun Li, Lewei Lu, Xizhou Zhu, et~al.
\newblock The all-seeing project v2: Towards general relation comprehension of the open world.
\newblock {\em arXiv preprint arXiv:2402.19474}, 2024.

\bibitem{wang2023allseeing}
Weiyun Wang, Min Shi, Qingyun Li, Wenhai Wang, Zhenhang Huang, Linjie Xing, Zhe Chen, Hao Li, Xizhou Zhu, Zhiguo Cao, et~al.
\newblock The all-seeing project: Towards panoptic visual recognition and understanding of the open world.
\newblock {\em arXiv preprint arXiv:2308.01907}, 2023.

\bibitem{wang2023beit3}
Wenhui Wang, Hangbo Bao, Li~Dong, Johan Bjorck, Zhiliang Peng, Qiang Liu, Kriti Aggarwal, Owais~Khan Mohammed, Saksham Singhal, Subhojit Som, et~al.
\newblock Image as a foreign language: Beit pretraining for vision and vision-language tasks.
\newblock In {\em CVPR}, 2023.

\bibitem{wang2022self}
Yizhong Wang, Yeganeh Kordi, Swaroop Mishra, Alisa Liu, Noah~A Smith, Daniel Khashabi, and Hannaneh Hajishirzi.
\newblock Self-instruct: Aligning language models with self-generated instructions.
\newblock {\em arXiv preprint arXiv:2212.10560}, 2022.

\bibitem{wang2022benchmarking}
Yizhong Wang, Swaroop Mishra, Pegah Alipoormolabashi, Yeganeh Kordi, Amirreza Mirzaei, Anjana Arunkumar, Arjun Ashok, Arut~Selvan Dhanasekaran, Atharva Naik, David Stap, et~al.
\newblock Benchmarking generalization via in-context instructions on 1,600+ language tasks.
\newblock {\em arXiv preprint arXiv:2204.07705}, 2, 2022.

\bibitem{wu2024visionllmv2}
Jiannan Wu, Muyan Zhong, Sen Xing, Zeqiang Lai, Zhaoyang Liu, Wenhai Wang, Zhe Chen, Xizhou Zhu, Lewei Lu, Tong Lu, et~al.
\newblock Visionllm v2: An end-to-end generalist multimodal large language model for hundreds of vision-language tasks.
\newblock {\em arXiv preprint arXiv:2406.08394}, 2024.

\bibitem{xu2024vision}
Zhiyang Xu, Chao Feng, Rulin Shao, Trevor Ashby, Ying Shen, Di~Jin, Yu~Cheng, Qifan Wang, and Lifu Huang.
\newblock Vision-flan: Scaling human-labeled tasks in visual instruction tuning.
\newblock {\em arXiv preprint arXiv:2402.11690}, 2024.

\bibitem{xu2022multiinstruct}
Zhiyang Xu, Ying Shen, and Lifu Huang.
\newblock Multiinstruct: Improving multi-modal zero-shot learning via instruction tuning.
\newblock {\em arXiv preprint arXiv:2212.10773}, 2022.

\bibitem{ye2023mplugdocowl}
Jiabo Ye, Anwen Hu, Haiyang Xu, Qinghao Ye, Ming Yan, Yuhao Dan, Chenlin Zhao, Guohai Xu, Chenliang Li, Junfeng Tian, Qian Qi, Ji~Zhang, and Fei Huang.
\newblock mplug-docowl: Modularized multimodal large language model for document understanding, 2023.

\bibitem{yin2024lamm}
Zhenfei Yin, Jiong Wang, Jianjian Cao, Zhelun Shi, Dingning Liu, Mukai Li, Xiaoshui Huang, Zhiyong Wang, Lu~Sheng, Lei Bai, et~al.
\newblock Lamm: Language-assisted multi-modal instruction-tuning dataset, framework, and benchmark.
\newblock {\em Advances in Neural Information Processing Systems}, 36, 2024.

\bibitem{yu2022coca}
Jiahui Yu, Zirui Wang, Vijay Vasudevan, Legg Yeung, Mojtaba Seyedhosseini, and Yonghui Wu.
\newblock Coca: Contrastive captioners are image-text foundation models.
\newblock {\em arXiv preprint arXiv:2205.01917}, 2022.

\bibitem{yu2023mmvet}
Weihao Yu, Zhengyuan Yang, Linjie Li, Jianfeng Wang, Kevin Lin, Zicheng Liu, Xinchao Wang, and Lijuan Wang.
\newblock Mm-vet: Evaluating large multimodal models for integrated capabilities.
\newblock {\em arXiv preprint arXiv:2308.02490}, 2023.

\bibitem{yue2023mmmu}
Xiang Yue, Yuansheng Ni, Kai Zhang, Tianyu Zheng, Ruoqi Liu, Ge~Zhang, Samuel Stevens, Dongfu Jiang, Weiming Ren, Yuxuan Sun, et~al.
\newblock Mmmu: A massive multi-discipline multimodal understanding and reasoning benchmark for expert agi.
\newblock {\em arXiv preprint arXiv:2311.16502}, 2023.

\bibitem{zhang2023llavar}
Yanzhe Zhang, Ruiyi Zhang, Jiuxiang Gu, Yufan Zhou, Nedim Lipka, Diyi Yang, and Tong Sun.
\newblock Llavar: Enhanced visual instruction tuning for text-rich image understanding.
\newblock {\em arXiv preprint arXiv:2306.17107}, 2023.

\bibitem{zhu2023minigpt4}
Deyao Zhu, Jun Chen, Xiaoqian Shen, Xiang Li, and Mohamed Elhoseiny.
\newblock Minigpt-4: Enhancing vision-language understanding with advanced large language models.
\newblock {\em arXiv preprint arXiv:2304.10592}, 2023.

\bibitem{zhu2022uni}
Jinguo Zhu, Xizhou Zhu, Wenhai Wang, Xiaohua Wang, Hongsheng Li, Xiaogang Wang, and Jifeng Dai.
\newblock Uni-perceiver-moe: Learning sparse generalist models with conditional moes.
\newblock {\em arXiv preprint arXiv:2206.04674}, 2022.

\bibitem{zhu2021uni}
Xizhou Zhu, Jinguo Zhu, Hao Li, Xiaoshi Wu, Xiaogang Wang, Hongsheng Li, Xiaohua Wang, and Jifeng Dai.
\newblock Uni-perceiver: Pre-training unified architecture for generic perception for zero-shot and few-shot tasks.
\newblock {\em arXiv preprint arXiv:2112.01522}, 2021.

\end{thebibliography}
